\documentclass{article}

\usepackage{PRIMEarxiv}

\usepackage[utf8]{inputenc} 
\usepackage[T1]{fontenc}    
\usepackage{hyperref}       
\usepackage{url}            
\usepackage{booktabs}       
\usepackage{amsfonts}       
\usepackage{nicefrac}       
\usepackage{microtype}      
\usepackage{lipsum}
\usepackage{graphicx}
\usepackage{amsmath, amssymb}
\usepackage{multirow}
\graphicspath{{media/}}     

\title{One-Shot Neural Architecture Search with Network Similarity Directed Initialization for Pathological Image Classification
}

\author{
  Renao Yan\\
  Shenzhen International Graduate School \\
  Tsinghua University \\
  China\\
  \texttt{yra21@mails.tsinghua.edu} \\
}

\begin{document}
\maketitle

\begin{abstract}
Deep learning-based pathological image analysis presents unique challenges due to the practical constraints of network design. Most existing methods apply computer vision models directly to medical tasks, neglecting the distinct characteristics of pathological images. This mismatch often leads to computational inefficiencies, particularly in edge-computing scenarios. To address this, we propose a novel Network Similarity Directed Initialization (NSDI) strategy to improve the stability of neural architecture search (NAS). Furthermore, we introduce domain adaptation into one-shot NAS to better handle variations in staining and semantic scale across pathology datasets. Experiments on the BRACS dataset demonstrate that our method outperforms existing approaches, delivering both superior classification performance and clinically relevant feature localization.
\end{abstract}


\section{Introduction}
Breast cancer is among the most prevalent and fatal malignancies affecting women worldwide, posing a major global health burden \cite{sung2021global}. Its metastatic nature—commonly spreading to the bones, liver, lungs, and brain—contributes significantly to its incurability \cite{sun2017risk,zhu2023accurate}. Early detection can dramatically increase the 5-year survival rate to 85\%, underscoring the importance of proactive screening \cite{ALLEMANI2015977}. However, manual examination of histological slides remains time-consuming and labor-intensive \cite{gomes2014inter,10.1001/jama.2015.1405,yan2024shapley,ouyang2024mergeup,wang2024task,zhu2025hierarchically}. With the rising incidence of breast cancer, there is a growing need for automated, efficient diagnostic tools. The digitization of pathology slides has enabled the production of gigapixel images, making large-scale computational analysis feasible.

Deep learning, particularly convolutional neural networks (CNNs), has greatly advanced pathological image analysis, offering improved diagnostic accuracy and efficiency. Nevertheless, designing architectures suited to pathology requires domain-specific considerations. Standard models from natural image domains—such as ResNet or InceptionNet—are often repurposed, but they fail to account for the simpler color distributions and richer hierarchical structures characteristic of pathological images. While researchers typically fine-tune these models using self-supervised or fully supervised approaches, such adaptations often overlook the unique demands of histopathology, resulting in suboptimal outcomes \cite{yan2023unpaired}. Moreover, clinical deployment requires not only accuracy but also computational efficiency when processing gigapixel whole slide images (WSIs) in a patch-wise manner on edge devices.

Neural architecture search (NAS), a key technique in automated machine learning (AutoML), has shown promise in generating efficient networks under constraints such as FLOPs or parameter count. Despite its potential, NAS remains underutilized in pathological imaging. Traditional NAS methods like evolutionary algorithms (EA) or Bayesian optimization (BO) often rely on random initialization (RI), which can lead to unstable performance when only a small number of samples are available due to high evaluation cost.

To address this, we propose a two-fold approach. First, we introduce a Network Similarity Directed Initialization (NSDI) strategy to enhance the stability and effectiveness of the search process. Second, we incorporate domain adaptation into one-shot NAS to address distributional shifts caused by varying staining protocols and semantic scales across pathology datasets. The inclusion of domain adaptation loss constrains the supernet to yield more reliable performance estimates, improving evaluation accuracy and overall search quality. Importantly, our method is modular and can be easily integrated into existing NAS pipelines.

This work makes the following key contributions:
\begin{itemize}
\item[$\bullet$] We propose a novel NSDI algorithm that improves the robustness of NAS by reducing redundancy in the initial population, particularly under limited initialization budgets.
\item[$\bullet$] We introduce domain adaptation into one-shot NAS using a Maximum Mean Discrepancy (MMD) loss to mitigate domain shifts in pathological datasets, enabling more reliable architecture evaluation.
\item[$\bullet$] Extensive experiments demonstrate that our method achieves superior classification performance and enhanced search stability compared to existing approaches.
\end{itemize}

\section{Related Works}
\subsection{Image Classification in Breast Cancer}
Most recent studies on breast cancer classification employ weakly supervised techniques, particularly multiple instance learning (MIL), to analyze whole slide image (WSI)-level data. Thandiackal \emph{et al.} proposed ZoomMIL, an end-to-end framework that performs multi-level zooming and outperforms state-of-the-art (SOTA) MIL approaches on the BRIGHT and CAMELYON16 datasets \cite{thandiackal2022differentiable}. Zhan \emph{et al.} integrated both region-of-interest (ROI) and WSI-level information for breast tumor classification in the BRIGHT Challenge \cite{zhan2022breast}. Wang \emph{et al.} introduced a weakly supervised method based on cross-slide contrastive learning, which decouples task-agnostic self-supervised feature extraction from task-specific feature refinement and aggregation \cite{wangscl}. Marini \emph{et al.} developed an instance-based MIL model that integrates both strongly and weakly labeled data through a multi-task loss \cite{marini2022multi}. Wentai \emph{et al.} proposed a MIL pipeline enhanced with transformers for subtype classification \cite{wentai2022multiple}.

Graph-based approaches have also gained attention. Hou \emph{et al.} designed a spatial-hierarchical graph neural network (GNN) with dynamic structure learning to model spatial dependencies \cite{hou2022spatial}. Pati \emph{et al.} introduced a hierarchical GNN that captures intra- and inter-entity interactions in tissue via entity graphs \cite{pati2022hierarchical}. Tiard \emph{et al.} proposed a self-supervised method that incorporates stain normalization into a constrained latent space for robust feature learning \cite{tiard2022stain}. Most of these methods focus on WSI-level classification, while patch-level classification remains less explored.

\subsection{Neural Architecture Search}
NAS typically consists of three components: the search space, the search strategy, and the evaluation method.

\subsubsection{Search Space}
Zoph \emph{et al.} initially proposed a flexible search space that optimizes the size, stride, and number of kernels in each convolutional layer \cite{zoph2016neural}. Later works such as NASNet \cite{zoph2018learning} introduced modular designs by stacking normal and reduction cells. Zhang \emph{et al.} explored block-based architectures \cite{zhang2020autobss}. Liu \emph{et al.} proposed DARTS, which searches over a continuous relaxation of the architecture space using a weight-sharing supernet \cite{liu2018darts}, significantly reducing search cost while achieving competitive results.

Another line of work focuses on optimizing existing architectures through search. ProxylessNAS \cite{cai2018proxylessnas}, MDENAS \cite{zheng2019multinomial}, MNasNet \cite{tan2019mnasnet}, FBNet \cite{wu2019fbnet}, and FBNetV2 \cite{wan2020fbnetv2} all extend MobileNetV2 by adjusting the number, size, and type of convolutional blocks.

\subsubsection{Search Strategy}
Popular search strategies include random search (RS), Bayesian optimization (BO), evolutionary algorithms (EA), reinforcement learning (RL), and gradient-based methods. Bergstra \emph{et al.} leveraged BO to identify optimal architectures. Zoph \emph{et al.} employed RL to navigate the search space \cite{zoph2016neural}, while Real \emph{et al.} demonstrated improved performance using a regularized EA \cite{real2019regularized}. Liu \emph{et al.} combined parameter sharing with gradient-based search in DARTS to further reduce computational cost \cite{liu2018darts}.

\subsubsection{Evaluation Strategy}
The standard evaluation approach involves training candidate architectures to convergence and assessing their performance on a validation set. Although accurate, this is computationally expensive. To reduce cost, Klein \emph{et al.} used partial training on subsets of data \cite{klein2017learning}, and Chrabaszcz \emph{et al.} used lower-resolution images \cite{chrabaszcz2017downsampled}. Domhan \emph{et al.} proposed early-stopping strategies based on performance extrapolation from early epochs \cite{domhan2015speeding}. Cai \emph{et al.} trained a surrogate model to predict architecture performance from structural encoding \cite{cai2018efficient}.

Recent one-shot NAS methods further reduce evaluation cost through weight-sharing supernets, where all candidate architectures share parameters and are evaluated without individual retraining \cite{guo2020single}.

\subsection{NAS Applications in Medical Image Analysis}
Despite its success in natural image tasks, NAS has seen limited use in medical imaging. Dong \emph{et al.} introduced a NAS framework for adversarial medical image segmentation \cite{dong2019neural}. Yan \emph{et al.} proposed MS-NAS, which fuses multi-scale features for cell-level tasks \cite{yan2020ms}. Tang \emph{et al.} applied a hyperparameter-tuned DARTS framework to computational pathology (CPath), achieving promising results on the ADP dataset \cite{tang2021probeable}. Huang \emph{et al.} developed AdwU-Net, a NAS framework that adapts the depth and width of U-Net for segmentation tasks in the Medical Segmentation Decathlon (MSD) \cite{huang2022adwu}. Eminaga \emph{et al.} proposed PlexusNet, a scalable model family tailored to five clinical classification tasks through structured control of depth, width, and branching \cite{eminaga2023plexusnet}.

\section{Method}

\begin{figure*}[h!]
\centering
\includegraphics[width=16cm]{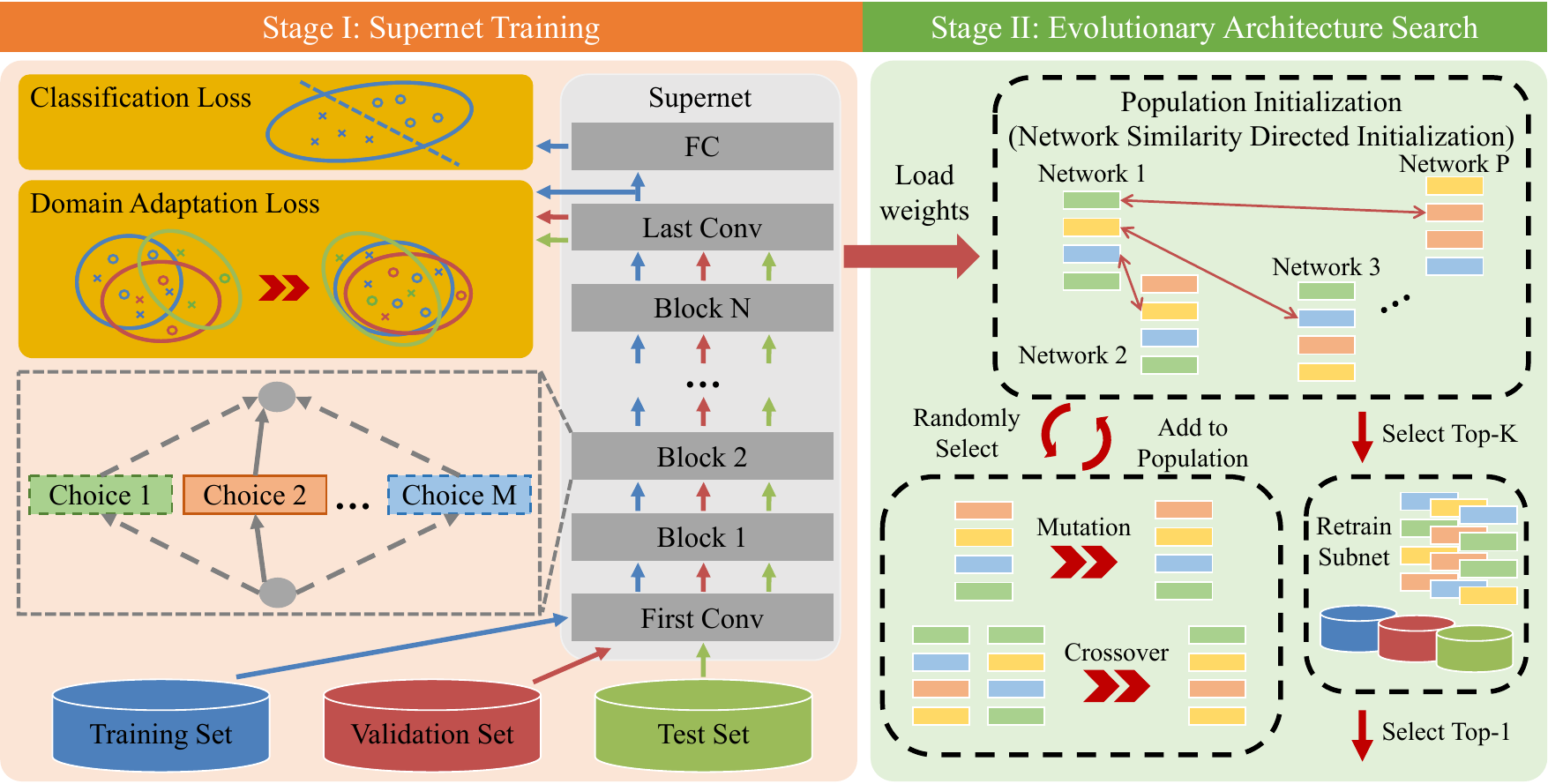}
\caption{The overview of the proposed domain adaptation one-shot neural architecture search.} \label{overview}
\end{figure*}

We propose a novel neural architecture search (NAS) framework tailored for pathological image analysis, termed Domain Adaptation One-Shot NAS (DAOS). It incorporates a network similarity directed initialization (NSDI) strategy to enhance search stability and introduces domain adaptation into the one-shot NAS paradigm. As shown in Fig.~\ref{overview}, the overall pipeline consists of two main stages: supernet training and architecture search.

The architecture search space $\mathcal A$ is modeled as a directed acyclic graph (DAG), where each candidate architecture corresponds to a subgraph $a \in \mathcal A$, denoted as $\mathcal N_{a,w}$ with weights $w$. In our design, the search space comprises $N$ layers, each offering $M$ candidate operations, yielding a total of $M^N$ possible architectures.

NAS typically involves training candidate architectures to convergence, then ranking them based on evaluation metrics such as accuracy or F1 score. This process is formalized as:
\begin{equation}
w_a = \mathop{\mathrm{argmin}} \limits_{w} \mathcal L_{\mathrm{train}}\left ( \mathcal N_{a,w}\right ),
\label{eq1}
\end{equation}
\begin{equation}
a^\ast = \mathop{\mathrm{argmax}} \limits_{a \in \mathcal A} \mathrm{Metrics}{\mathrm{val}}\left ( \mathcal N{a,w}\right ),
\label{eq2}
\end{equation}
where $\mathcal{L}{\mathrm{train}}(\cdot)$ is the training loss, and $\mathrm{Metrics}{\mathrm{val}}(\cdot)$ denotes validation performance.

\subsection{Search Algorithm}

\begin{figure}[h!]
\centering
\includegraphics[width=10cm]{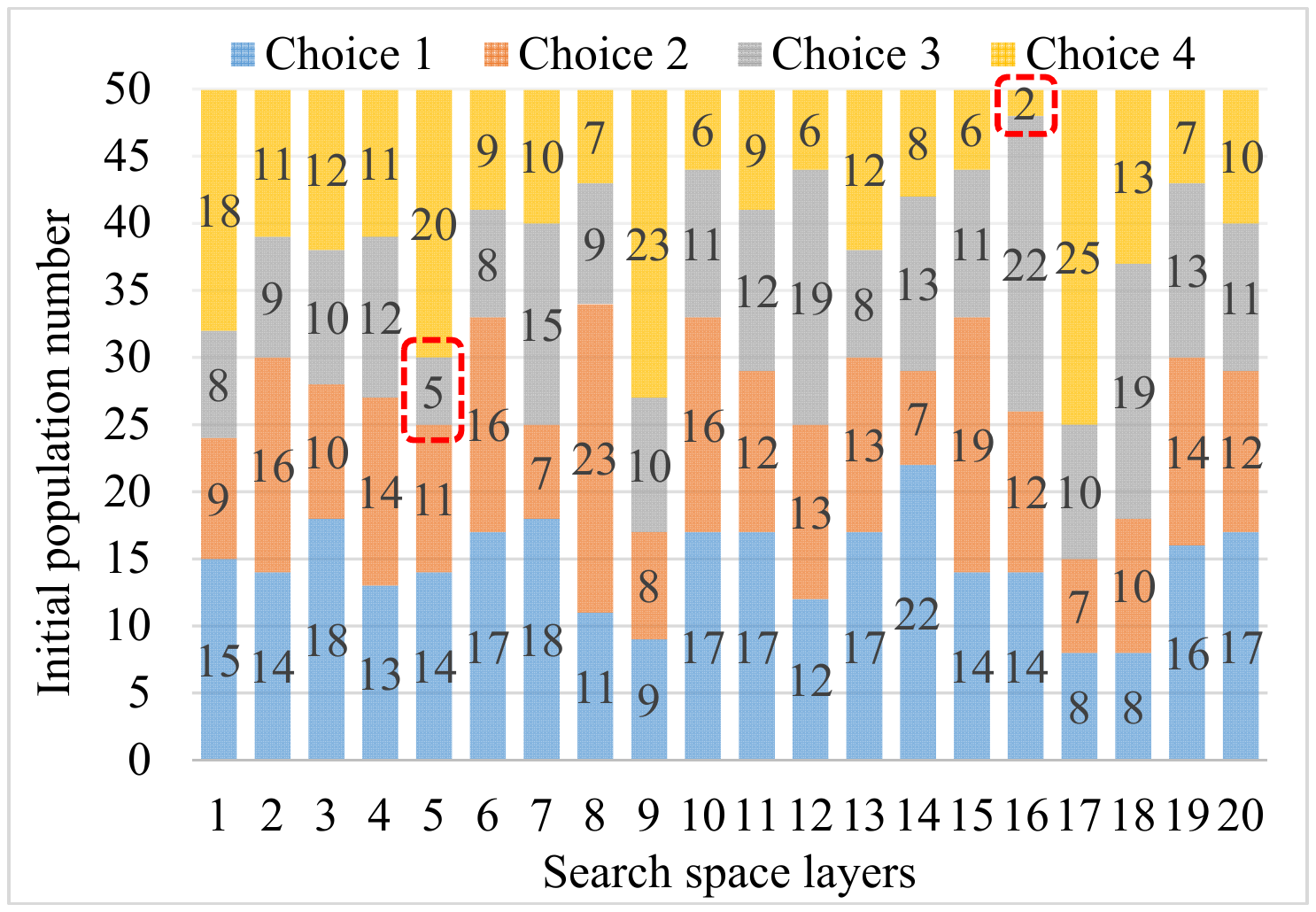}
\caption{One sample of population initialization under a specific random seed, where the population number is 50, and the search space is a 20-layer network, with four choices in each layer.} \label{RI}
\end{figure}

We adopt an evolutionary algorithm (EA) for architecture search, using random initialization (RI) to generate the initial population. However, when the population size is small, pseudo-random sampling can result in an imbalanced exploration of the search space. Fig.~\ref{RI} shows that RI often fails to evenly sample operation choices (e.g., the red box highlights under-represented operators). Although methods like AutoBSS \cite{zhang2020autobss} partially address this using clustering-based initialization, they lack consistency.

To improve population diversity and search stability, we introduce a Network Similarity Directed Initialization (NSDI) method, inspired by force-directed placement (FDP) algorithms \cite{fruchterman1991graph}. FDP arranges nodes based on repulsive and attractive forces but operates in continuous space, making it unsuitable for discrete search spaces like NAS. Instead, we define a discrete similarity metric to guide initialization.

Specifically, the similarity between two network samples is defined as:
\begin{equation}
\mathrm{SS}(v_i,v_j) \triangleq \sum_{k=0}^{N} v_{i,k} \odot v_{j,k},
\label{eq9}
\end{equation}
where $v_i$ and $v_j$ are $N$-dimensional binary vectors encoding architectures $a_i$ and $a_j$, and $\odot$ denotes the XNOR operation.

We then define the Average Population Similarity (APS) as:
\begin{equation}
\mathrm{APS}(V^P) \triangleq \frac{1}{P} \sum_{i=0}^{P} \max_{v_j \in V^P, i \neq j} \mathrm{SS}(v_i,v_j),
\label{eq10}
\end{equation}
where $V^P$ is the set of $P$ encoded network vectors.

To ensure diversity, we constrain APS under a user-defined threshold $APS_{max}$ and increment it adaptively to avoid excessive sampling time. The NSDI process is summarized in Algorithm~1.

\begin{table}[htbp]
\centering
\renewcommand\arraystretch{1.2}
\begin{tabular}{p{0.5cm} p{13cm}}  
\toprule
\multicolumn{2}{l}{\textbf{Algorithm 1:} Network Similarity Directed Initialization} \\
\midrule
1  & \textbf{Input:} population size $P$, search depth $N$, $M$ choices per layer, latency bound $Lat_{max}$, similarity threshold $APS_{max}$, timeout $T$ \\
2  & \textbf{Output:} initialized population $V$ \\
3  & While len($V$) $< P$: \\
4  & \quad $v$ = randint($M$, $N$) \\
5  & \quad if len($V$) == 0: $V \leftarrow v$ \\
6  & \quad else: \\
7  & \qquad $SS_{\max} = \max_{u \in V} \mathrm{SS}(v, u)$ \\
8  & \qquad if $SS_{\max} \leq APS_{max}$ and Latency($v$) $\leq Lat_{max}$: \\
9  & \qquad\qquad $V \leftarrow v$ \\
10 & \qquad else: $t \leftarrow t + 1$ \\
11 & \quad if $t > T$: $APS_{max} \leftarrow APS_{max} + 1$; $t \leftarrow 0$ \\
12 & Return population $V$ \\
\bottomrule
\end{tabular}
\end{table}

\subsection{Supernet Training}

To avoid retraining every architecture, we adopt a one-shot NAS approach using a shared-weight supernet $\mathcal N_{\mathcal A, W}$, which spans the entire search space $\mathcal A$ with shared weights $W$. After training, candidate architectures inherit weights from the supernet and are evaluated directly, significantly reducing search cost.

The supernet weights are optimized by minimizing:
\begin{equation}
W_\mathcal A = \mathop{\mathrm{argmin}}W \mathcal L{\mathrm{train}}(\mathcal N_{\mathcal A, W}),
\label{eq3}
\end{equation}
often approximated by sampling architectures from a prior $\Gamma(\mathcal A)$:
\begin{equation}
W_\mathcal A = \mathop{\mathrm{argmin}}W \mathbb{E}{a \sim \Gamma(\mathcal A)} [\mathcal L_{\mathrm{train}}(\mathcal N_{\mathcal A, W})],
\label{eq4}
\end{equation}
where $\Gamma(\mathcal A)$ is uniformly sampled under FLOPs constraints.

Given a labeled training set $\mathcal D_s = {(x_i^s, y_i^s)}{i=1}^{n_s}$, where $y_i^s \in \mathbb{R}^C$ is a one-hot label, the classification loss is:
\begin{equation}
\mathcal L{\mathrm{cls}}(\mathcal N_{\mathcal A, W}, \mathcal D_s) = \frac{1}{n_s} \sum_{i=1}^{n_s} J(\mathcal N_{a, W_\mathcal A(a)}(x_i^s), y_i^s),
\label{eq5}
\end{equation}
where $J(\cdot,\cdot)$ denotes cross-entropy loss.

\subsection{Domain Adaptation with MMD}

Weight sharing can lead to performance estimation bias, especially when the training and validation distributions differ (e.g., due to stain variability). To mitigate this, we introduce a domain adaptation loss using Maximum Mean Discrepancy (MMD).

Assume source domain $\mathcal D_s$ (labeled training) and target domain $\mathcal D_t = {x_j^t}{j=1}^{n_t}$ (unlabeled validation), sampled from different distributions $p \neq q$. MMD quantifies their discrepancy as:
\begin{equation}
d\mathcal H(p, q) \triangleq \left| \mathbb{E}_p[\phi(x^s)] - \mathbb{E}q[\phi(x^t)] \right|^2\mathcal H,
\label{eq6}
\end{equation}
where $\phi(\cdot)$ maps samples to a reproducing kernel Hilbert space (RKHS) with kernel $k(x, x’) = \langle \phi(x), \phi(x’) \rangle$.

An unbiased estimator of MMD is:
\begin{equation}
\begin{aligned}
\hat{d}\mathcal H(p, q) &= \frac{1}{n_s^2} \sum{i,j} k(x_i^s, x_j^s) + \frac{1}{n_t^2} \sum_{i,j} k(x_i^t, x_j^t) \
&\quad - \frac{2}{n_s n_t} \sum_{i,j} k(x_i^s, x_j^t).
\end{aligned}
\label{eq7}
\end{equation}

The final training objective combines classification and domain adaptation loss:
\begin{equation}
\mathcal L_{\mathrm{train}}(\mathcal N_{\mathcal A, W}) = \mathcal L_{\mathrm{cls}} + \lambda \sum_{\gamma \in {q,l}} \hat{d}_\mathcal H(p, \gamma),
\label{eq8}
\end{equation}
where $\lambda$ is a balancing coefficient (set to 0.5 in our experiments).

\section{Experiment}

\subsection{Dataset}

Due to the high computational cost of neural architecture search (NAS), we focus our evaluation on a single large-scale dataset—BRACS \cite{brancati2022bracs}. The dataset comprises 4,391 breast histological images, scanned using an Aperio AT2 scanner at a resolution of 0.25 $\mu$m/pixel. All tumor regions-of-interest (TRoIs) are annotated into eight categories: Normal, Benign, Usual Ductal Hyperplasia (UDH), Atypical Ductal Hyperplasia (ADH), Flat Epithelial Atypia (FEA), Ductal Carcinoma In Situ (DCIS), and Invasive. TRoI images vary in spatial resolution, with an average size of 1778$\times$1723 pixels.

\subsection{Training Details}

We follow the data augmentation protocol introduced in \cite{touvron21a}, resizing input images to 512$\times$512 pixels. All models are trained using the Adam optimizer ($\beta_1$=0.9, $\beta_2$=0.999, $\epsilon$=1e-8) to minimize cross-entropy loss. Training is performed on NVIDIA Tesla V100 GPUs using PyTorch v1.7.1 with a batch size of 64.

\subsubsection{Search Space}

We adopt the one-shot ShuffleNet V2-based search space from \cite{guo2020single}. The search space consists of $N=20$ block layers, each offering $M=4$ candidate operations. These include Shuffle blocks with kernel sizes of 3×3, 5×5, and 7×7, as well as Xception-style blocks with varying depthwise convolutions. The resulting search space contains $4^{20}$ possible architectures.

\subsubsection{Supernet Training}

The supernet is initialized with pre-trained weights from ImageNet \cite{guo2020single} and trained for 2000 epochs on BRACS using the single-path strategy as baseline. In the DAOS-A variant, the supernet is further fine-tuned for the final 1000 epochs using a combination of classification and domain adaptation losses. In DAOS-B, the classifier is frozen and the encoder is further fine-tuned for an additional 1000 epochs. The initial learning rate is set to 3e-4, and the fine-tuning rate to 1.5e-4. A cosine annealing schedule reduces the learning rate to a minimum of 6e-7.

\subsubsection{Search algorithm}We use evolutionary algorithm with the initialization population $P=100$, among which the top 50 candidates are selected as the population for further EA-based search. For mutation, a randomly selected candidate mutates its every choice block with a probability of 0.1 to produce a new candidate. For crossover, two randomly selected candidates are crossed to produce a new one. Mutation and crossover are repeated (every 25 operations) for enough new candidates. FLOPs $\leq1800$M is adopted as the complexity constraint because of the large image size. A total of 1000 candidates are search and the top 10 networks are selected for retraining.
\subsubsection{Candidate Retraining}

Top-ranked architectures from the search stage inherit weights from the supernet and are then fine-tuned on BRACS for 50 epochs. The learning rate is scheduled from 5e-6 to 3e-4 using triangular annealing.

\subsection{Results and Analysis}

\subsubsection{NSDI Enhances Search Stability}

NAS is inherently a black-box optimization problem, where candidate architectures are encoded as discrete vectors within a combinatorial search space. Due to the high cost of evaluating each network, initializing the population with a representative and diverse set is critical to guiding the search effectively.

\begin{figure}
\centering
\includegraphics[width=9cm]{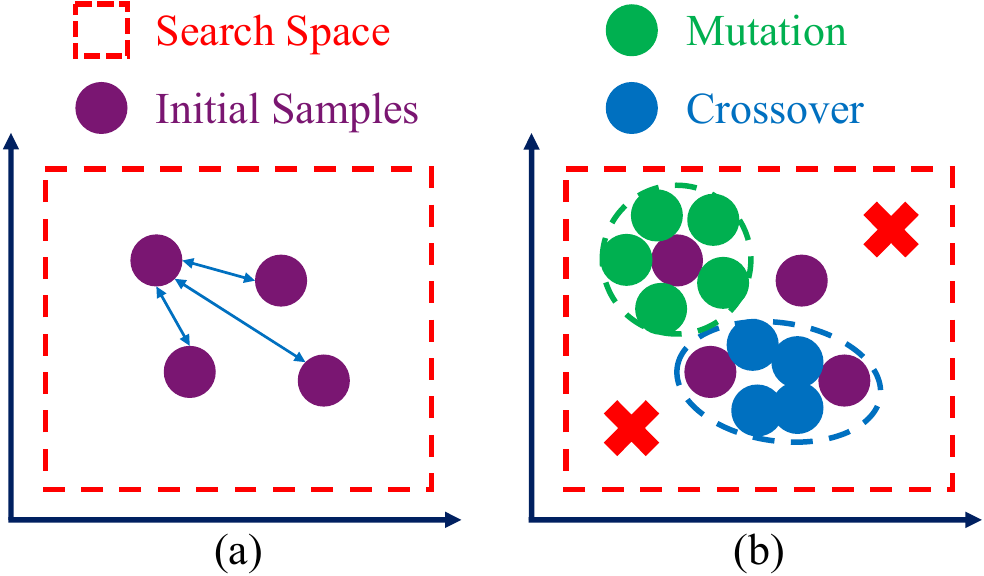}
\caption{The impact of population initialization on the evolutionary algorithm. (a)Random sampling, (b)mutation and crossover can fall into local optimum.} \label{EA}
\end{figure}

As illustrated in Fig.\ref{EA}, evolutionary algorithms (EA) rely on mutation and crossover, which are heavily influenced by the initial population. Poor initialization can trap the search in local optima. To demonstrate this, we compute the average population similarity (APS) for a random-initialized population with $P=50$, $N=20$, $M=4$, and $Lat_{max}=1800$M FLOPs (Table\ref{table1}). Results show that random initialization yields populations where each architecture is, on average, 50\% similar to at least one other sample.
\begin{table}[]
\caption{Influence of different population initialization methods on average population similarity}
\centering
\renewcommand\arraystretch{1.2}
\begin{tabular}{lllllll}
\toprule
     & Random Seed & 1     & 2    & 3    & 4    & 5     \\ \hline
RI   & APS         & 10.06 & 9.94 & 9.76 & 9.8  & 10.22 \\
NSDI & APS         & 6.98  & 7.00 & 7.00 & 6.98 & 6.98  \\
\bottomrule
\end{tabular}
\label{table1}
\end{table}

By contrast, our NSDI strategy—with $APS_{max}=6$ and timeout $T=2\times10^5$—achieves lower APS scores, leading to better diversity, albeit at the cost of more sampling. This trade-off is manageable and can be tuned via $T$.

\begin{figure}[h!]
\centering
\includegraphics[width=13cm]{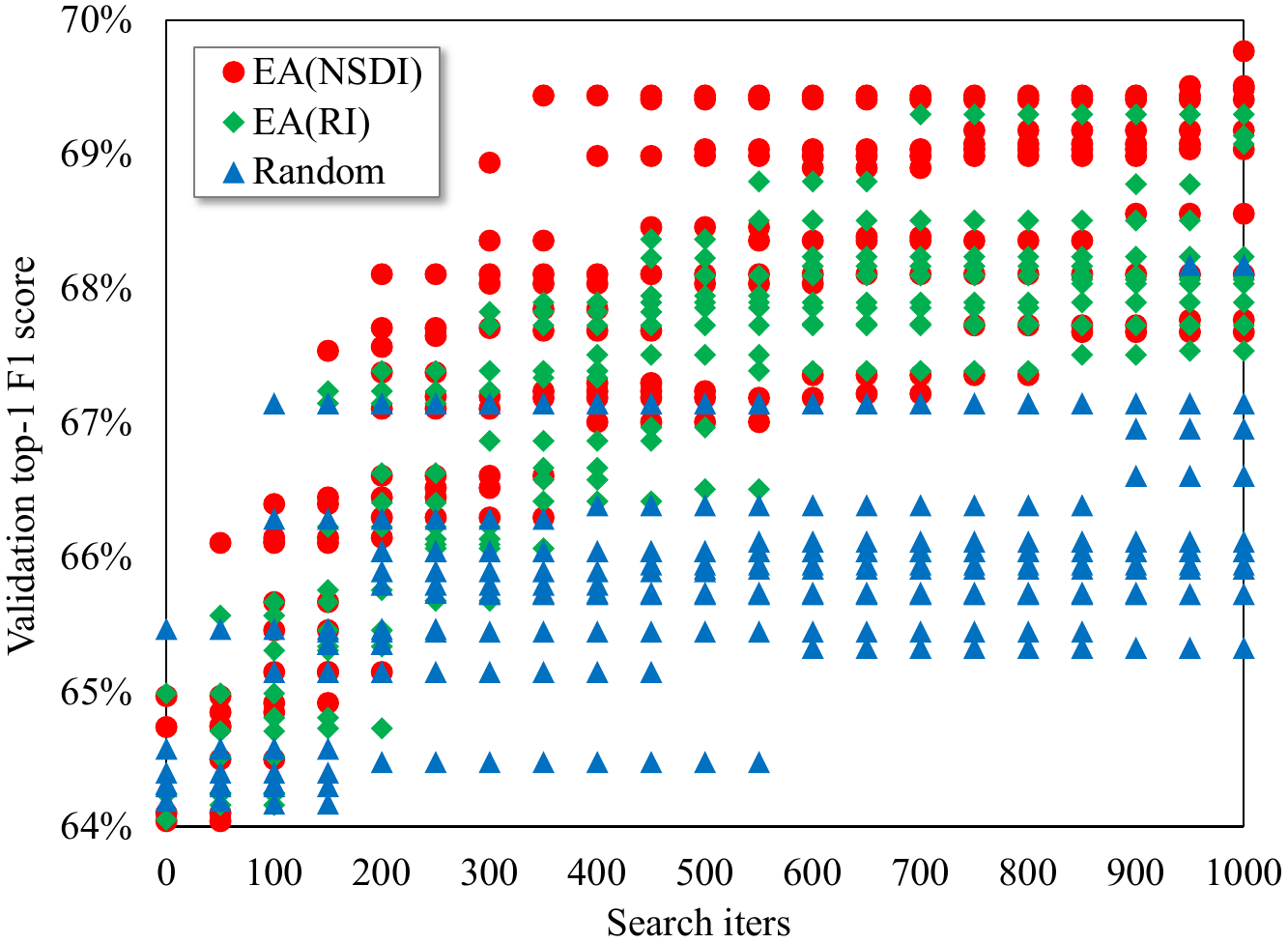}
\caption{Comparison of random search and evolutionary algorithm with random initialization or network similarity directed initialization.} \label{process}
\end{figure}

Fig.\ref{process} compares the F1 score trends across generations for three search strategies. Each method is repeated 10 times. 

Table\ref{table5} summarizes the best performance per trial, showing that EA with NSDI consistently outperforms both random search and EA with random initialization. Despite the time-intensive nature of NAS, our method ensures that the result of each search is consistently close to the global optimum.
\begin{table}[h!]
\caption{Best search results on validation set}
\centering
\renewcommand\arraystretch{1.2}
\label{table5}
\begin{tabular}{llll}
\toprule
Seed & Random & EA(RI) & EA(NSDI) \\ \hline
1 & 65.73 & 69.66 & 69.77 \\
2 & 65.33 & 68.45 & 69.04 \\
3 & 66.61 & 67.68 & 68.56 \\
4 & 68.18 & 68.92 & 67.68 \\
5 & 66.96 & 67.59 & 69.41 \\
6 & 66.05 & 67.63 & 69.51 \\
7 & 67.15 & 68.62 & 69.49 \\
8 & 66.12 & 67.30 & 68.09 \\
9 & 65.95 & 69.72 & 69.19 \\
10 & 65.92 & 68.13 & 68.11 \\ \hline
Mean & 66.40 & 68.37 & 68.89 \\
Std & 0.79 & 0.82 & 0.69
\end{tabular}
\end{table}

In terms of search cost, DAOS-A matches the training time of SPOS (Table~\ref{table4}). DAOS-B incurs slightly more cost due to the additional fine-tuning of the encoder.

\begin{table}[h!]
\caption{Search cost (GPU hours - Ghs)}
\centering
\renewcommand\arraystretch{1.2}
\begin{tabular}{llll}
\toprule
Method        & SPOS   & DAOS-A & DAOS-B \\ \hline
Training time (8 GPUs in total) & 48 Ghs & 48 Ghs & 72 Ghs \\
Search time   & 16 Ghs & 16 Ghs & 16 Ghs \\
Retrain time  & 8 Ghs  & 8 Ghs  & 8 Ghs  \\
Total time    & 72 Ghs & 72 Ghs & 96 Ghs \\
\bottomrule
\end{tabular}
\label{table4}
\end{table}

\begin{figure}[h!]
\centering
\includegraphics[width=7cm]{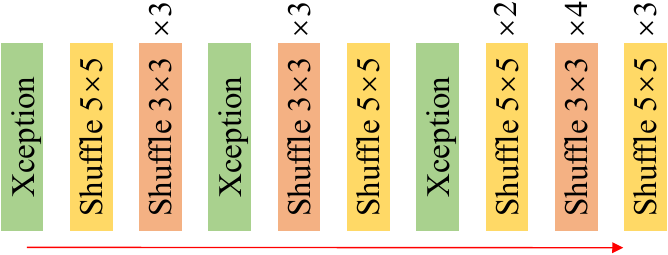}
\caption{Best architecture searched by DAOS.} \label{best}
\end{figure}

After 1000 search iterations, the top-10 architectures are retrained for evaluation. The best-discovered architecture (F1 = 61.41\%) is shown in Fig.~\ref{best} and the final performance results are presented in Table~\ref{table2}. Although random search remains a strong baseline, SPOS and EA+NSDI consistently outperform it. Both DAOS-A and DAOS-B further enhance performance and stability, with DAOS-B yielding the most stable outcomes.

\begin{table*}[h!]
\caption{Top-1 F1 score (\%) results on BRACS dataset with different neural architecture search methods (CL: Classification Loss, DAL: Domain Adaptation Loss, FE: Finetune Encoder)}
\centering
\renewcommand\arraystretch{1.2}
\resizebox{\textwidth}{!}{
\begin{tabular}{ccccccccc}
\toprule
                                                                & \multicolumn{2}{c}{Random}              & \multicolumn{2}{c}{SPOS}                    & \multicolumn{2}{c}{DAOS-A}                                                                                 & \multicolumn{2}{c}{DAOS-B}                                                                                                      \\ \hline
\begin{tabular}[c]{@{}c@{}}Supernet\\ Training\end{tabular}     & \multicolumn{2}{c}{CL} & \multicolumn{2}{c}{CL}     & \multicolumn{2}{c}{CL + DAL} & \multicolumn{2}{c}{CL + DAL + FE} \\ \hline
Search Method                                                   & \multicolumn{2}{c}{Random}              & \multicolumn{2}{c}{EA} & \multicolumn{2}{c}{EA}                                                                & \multicolumn{2}{c}{EA}                                                                                     \\ \hline
\begin{tabular}[c]{@{}c@{}}Initialization\\ Method\end{tabular} & \multicolumn{2}{c}{Random}              & \multicolumn{2}{c}{Random}                  & \multicolumn{2}{c}{NSDI}                                                                                   & \multicolumn{2}{c}{NSDI}                                                                                                        \\ \hline
Random Seed                                                     & FLOPs/Params      & Top-1 F1            & FLOPs/Params        & Top-1 F1              & FLOPs/Params                                        & Top-1 F1                                             & FLOPs/Params                                                  & Top-1 F1                                                        \\
1                                                               & 1716M/3.51M       & 58.96               & 1727M/3.46M         & 58.86                 & 1715M/3.20M                                         & 58.54                                                & 1727M/3.31M                                                   & 60.08                                                           \\
2                                                               & 1718M/3.34M       & 58.50               & 1761M/3.22M         & 57.87                 & 1736M/3.23M                                         & 59.33                                                & 1673M/3.22M                                                   & \textbf{60.45}                                                  \\
3                                                               & 1742M/3.34M       & 58.76               & 1707M/3.37M         & 58.17                 & 1713M/3.24M                                         & 58.37                                                & 1715M/3.28M                                                   & 58.52                                                           \\
4                                                               & 1745M/3.36M       & 58.60               & 1708M/3.44M         & 58.23                 & 1686M/3.22M                                         & 59.26                                                & 1735M/3.22M                                                   & 59.56                                                           \\
5                                                               & 1758M/3.42M       & 58.59               & 1741M/3.38M         & 57.36                 & 1683M/3.20M                                         & 58.98                                                & 1764M/3.22M                                                   & 57.71                                                           \\
6                                                               & 1719M/3.33M       & 58.70               & 1730M/3.38M         & 57.76                 & 1716M/3.19M                                         & 58.72                                                & 1681M/3.21M                                                   & 59.47                                                           \\
7                                                               & 1703M/3.48M       & 57.89               & 1738M/3.25M         & 59.21                 & 1706M/3.21M                                         & 59.61                                                & 1736M/3.21M                                                   & 59.56                                                           \\
8                                                               & 1763M/3.23M       & \textbf{59.27}      & 1631M/3.32M         & \textbf{59.54}        & 1716M/3.22M                                         & \textbf{61.41}                                       & 1653M/3.23M                                                   & 59.63                                                           \\
9                                                               & 1685M/3.37M       & 57.58               & 1722M/3.29M         & 58.58                 & 1689M/3.22M                                         & 60.51                                                & 1755M/3.32M                                                   & 59.45                                                           \\
10                                                              & 1739M/3.24M       & 58.29               & 1742M/3.32M         & 58.73                 & 1658M/3.21M                                         & 58.05                                                & 1733M/3.20M                                                   & 59.06                                                           \\ \hline
Best                                                            & 1763M/3.23M       & 59.27               & 1631M/3.32M         & 59.54                 & 1716M/3.22M                                         & \textbf{61.41}                                                & 1673M/3.22M                                                   & 60.45                                                           \\
Average                                                         & 1729M/3.36M       & 58.51$\pm$0.47          & 1721M/3.34M         & 58.43$\pm$0.64            & 1702M/3.21M                                         & 59.28$\pm$0.97                                           & 1717M/3.24M                                                   & \textbf{59.35$\pm$0.73}
\\
\bottomrule
\end{tabular}
}
\label{table2}
\end{table*}

\subsubsection{Domain Adaptation Improves Supernet Training}

\begin{figure}[h!]
\centering
\includegraphics[width=10cm]{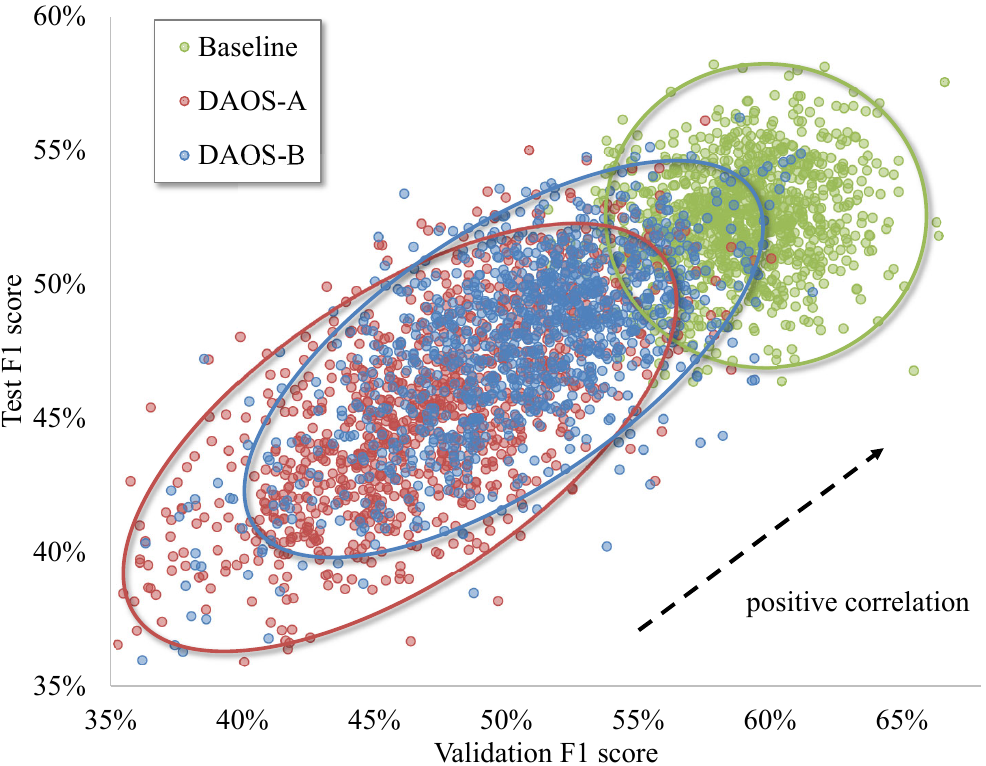}
\caption{Results of 1000 network evaluation metrics on the validation set and test set. All networks were randomly sampled from the supernet under different training methods. The baseline only uses the classification loss on the training set. DAOS-A uses classification loss and domain adaptation loss. DAOS-B fine-tunes the feature extractor based on DAOS-A by freezing the classifier.} \label{val2test}
\end{figure}

We compare three supernet training schemes:

\begin{itemize}
\item[$\bullet$] \textbf{Baseline:} 2000 epochs of training using classification loss only.
\item[$\bullet$] \textbf{DAOS-A:} Initial 2000 epochs with classification loss, followed by 1000 epochs with added domain adaptation loss.
\item[$\bullet$] \textbf{DAOS-B:} Further fine-tuning of the encoder for 1000 epochs while freezing the classifier (after DAOS-A).
\end{itemize}
To assess the effect of domain adaptation, we randomly sample 1000 architectures and evaluate their F1 scores on both validation and test sets. As shown in Fig.~\ref{val2test}, the baseline model shows poor correlation between validation and test metrics, meaning a model that performs well on the validation set may generalize poorly. This undermines the reliability of one-shot NAS.

With domain adaptation, both DAOS-A and DAOS-B significantly improve this consistency, yielding a more monotonic relationship between validation and test F1 scores. DAOS-B exhibits the strongest correlation.

Pearson correlation coefficients are computed to quantify this alignment: 0.1794 (Baseline), 0.6985 (DAOS-A), and 0.7096 (DAOS-B). This confirms the effectiveness of our domain adaptation design in stabilizing performance estimation.
\begin{figure*}
\centering
\includegraphics[width=\textwidth]{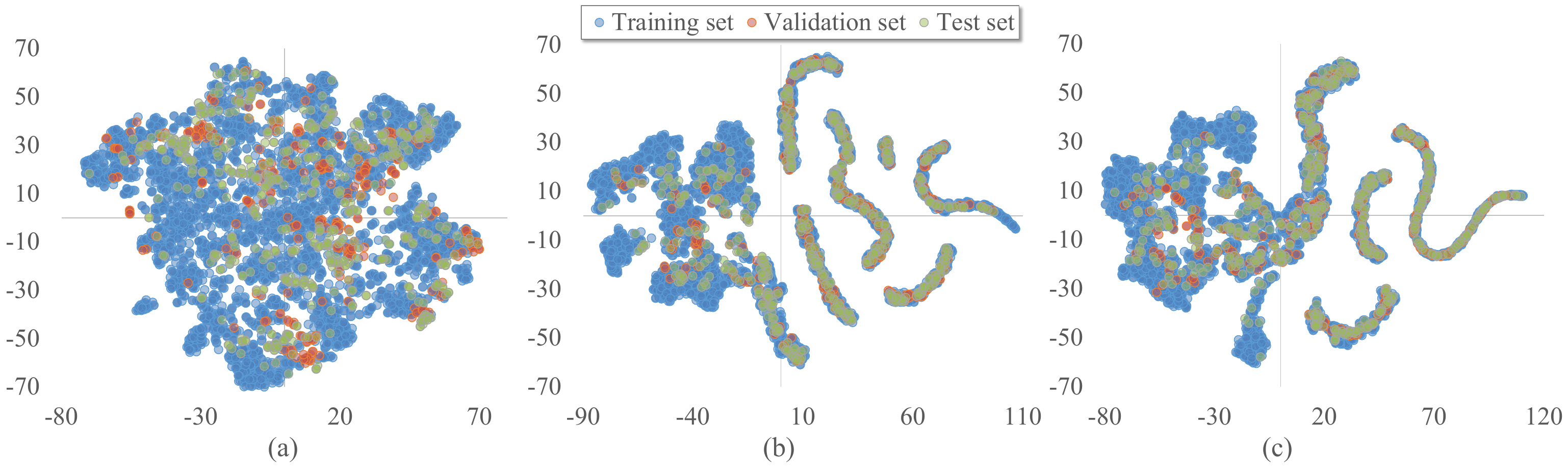}
\caption{t-SNE of the average supernet features on the training set by different methods. (a) Baseline, (b) DAOS-A, and (c) DAOS-B.} \label{tsne}
\end{figure*}
We further visualize the average feature representations of 1000 networks using t-SNE (Fig.~\ref{tsne}). Features from the baseline supernet are scattered and noisy, whereas those from DAOS-A and DAOS-B exhibit clearer separation and clustering, suggesting improved feature consistency.

\subsubsection{DAOS Achieves Superior Performance}

To benchmark overall performance, we compare our method against manually designed networks (e.g., ResNet18, InceptionNet V3, MobileNet V2, SqueezeNet variants) and NAS-based models. All models are initialized with ImageNet pre-trained weights and trained on BRACS for 100 epochs with a learning rate of 3e-4, using the same training configuration as for retraining searched candidates.

Results are reported in Table~\ref{table3}. DAOS-A achieves the highest F1 score among all methods under comparable FLOPs and parameter constraints, demonstrating its efficacy in constrained medical settings.
\begin{table}[h!]
\caption{F1 score (\%) results on BRACS dataset with different deep learning methods}
\centering
\renewcommand\arraystretch{1.2}
\begin{tabular}{cccc}
\toprule
                                                                                        & Method          & FLOPS/Params  & F1 score \\ \hline
\multirow{5}{*}{\begin{tabular}[c]{@{}c@{}}Manually\\ Designed\\ Networks\end{tabular}} & ResNet18        & 9527M/11.18M  & 60.28    \\
                                                                                        & InceptionNet V3 & 17629M/21.80M & 58.46    \\
                                                                                        & MobileNet V2    & 1704M/2.23M   & 57.96    \\
                                                                                        & SqueezeNet V1.0 & 3989M/0.74M   & 53.47    \\
                                                                                        & SqueezeNet V1.1 & 1446M/0.73M   & 56.56    \\ \hline
\multirow{4}{*}{\begin{tabular}[c]{@{}c@{}}Neural\\ Architecture\\ Search\end{tabular}} & Random Search   & 1731M/3.37M   & 59.27    \\
                                                                                        & SPOS            & 1721M/3.34M   & 59.44    \\
                                                                                        & DAOS-A          & 1702M/3.21M   & 61.41    \\
                                                                                        & DAOS-B          & 1717M/3.24M   & 60.45    \\
\bottomrule
\end{tabular}
\label{table3}
\end{table}
\subsubsection{DAOS Improves Pathological Interpretability}

Beyond accuracy, interpretability is crucial in clinical applications. We visualize Class Activation Maps (CAMs) \cite{zhou2016learning} for several patches using ResNet18, MobileNet V2, and our method.

\begin{figure*}
\centering
\includegraphics[width=\textwidth]{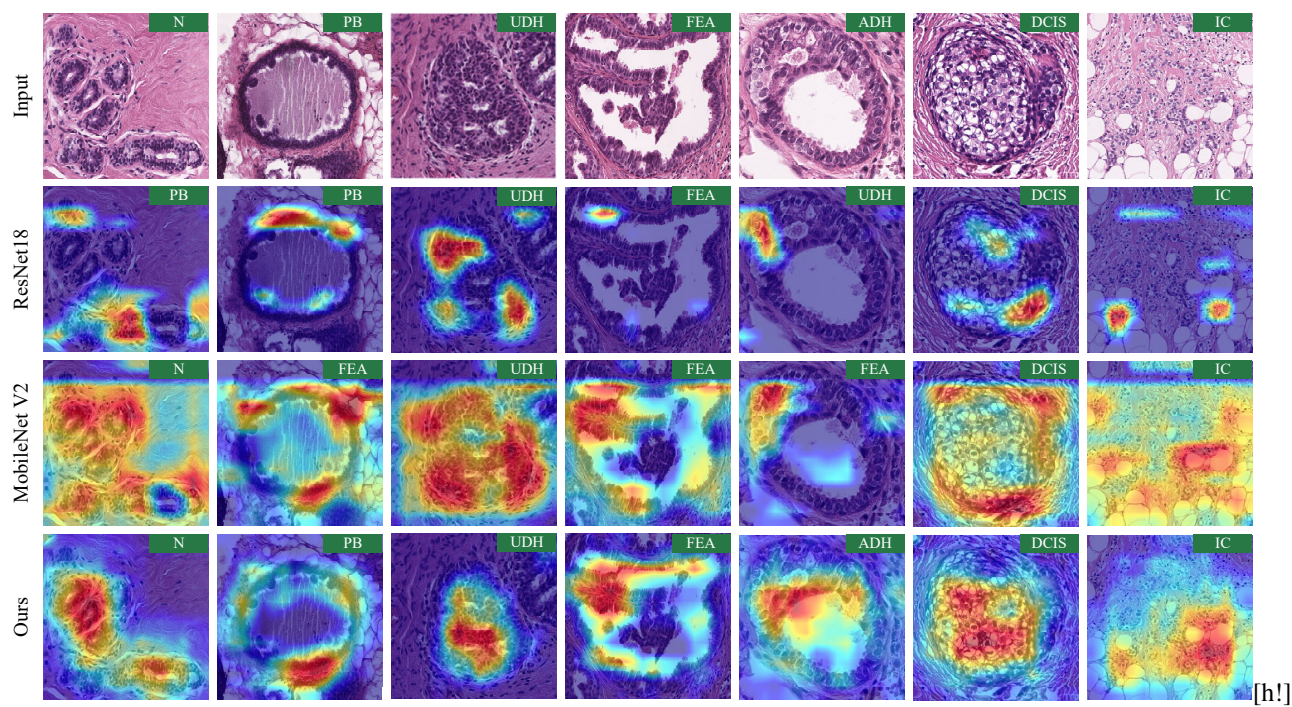}[h!]
\caption{CAMs generated from different methods. The upper left corner of the images show the real labels or predicted labels.} \label{heatmap}
\end{figure*}

As shown in Fig.~\ref{heatmap}, ResNet18 highlights limited tumor and epithelial regions, while MobileNet V2 attends to irrelevant areas like fat or connective tissue. In contrast, DAOS focuses on a wide range of diagnostically relevant regions—including tumor cells, ducts, and proliferative epithelial structures—better supporting subtype classification. This indicates that our method not only improves accuracy but also enhances clinical relevance and decision support.

\section{Conclusion}

In this work, we revisit the one-shot neural architecture search (NAS) paradigm in the context of medical image analysis and identify key limitations when applied to pathological datasets. To address these challenges, we propose a novel framework—Domain Adaptation One-Shot NAS (DAOS)—which integrates a network similarity directed initialization (NSDI) algorithm to improve search stability and introduces domain adaptation to enhance evaluation consistency under domain shifts.

To the best of our knowledge, this is the first work that incorporates domain adaptation into one-shot NAS and explicitly quantifies population diversity during initialization. Our approach achieves strong correlation between validation and test performance, enabling more reliable architecture ranking and selection.

Extensive experiments on the BRACS dataset demonstrate that both DAOS-A and DAOS-B consistently identify high-performing architectures close to the global optimum, outperforming existing baselines in terms of both accuracy and stability. Moreover, the proposed method is highly modular and can be readily extended to other NAS settings, including those based on graph neural networks or Bayesian optimization.

\bibliographystyle{unsrt}  
\bibliography{references}

\end{document}